\documentclass[12pt, draftclsnofoot, onecolumn]{IEEEtran}
\usepackage{amssymb}
\usepackage{amsfonts}
\usepackage{bbm}
\usepackage{amsfonts}
\usepackage[dvips]{graphicx}
\usepackage[dvips]{color}
\usepackage{graphicx}
\usepackage{amsmath}
\usepackage{fancyhdr}
\usepackage{stfloats}
\usepackage{multirow}
\usepackage{algorithm}
\usepackage{algorithmic}
\usepackage{cite}
\usepackage[bookmarks=false]{}
\usepackage{amsthm}
\usepackage{algorithm}
\usepackage{algorithmic}
\usepackage{mathbbol}
\usepackage{amsfonts}
\usepackage{graphicx}
\usepackage{amsmath}
\usepackage{amssymb}
\usepackage{latexsym}
\usepackage{graphicx}
\usepackage{cite}
\usepackage{url}
\usepackage{stfloats}
\usepackage{mathrsfs}
\usepackage{setspace}
\usepackage{booktabs}
\usepackage{latexsym}
\usepackage{url}
\usepackage{stfloats}
\usepackage{mathrsfs}
\theoremstyle{plain}

\usepackage{cite}
\usepackage{setspace}
\usepackage{textcomp}
\usepackage{xcolor}
\usepackage{makecell}
\usepackage{etoolbox}

\usepackage[tight,footnotesize]{subfigure}
\makeatletter
\renewcommand{\maketag@@@}[1]{\hbox{\m@th\normalsize\normalfont#1}}%
\makeatother

\begin{document}
%\pagenumbering{gobble}% Remove page numbers (and reset to 1)
\clearpage
%\maketitle
%Title.
% ------
\title{\huge Semantic Communications With AI Tasks}
%Beijing Key Laboratory of Network System Architecture and Convergence, School of Information and Communication Engineering, Beijing University of Posts and Telecommunications, Beijing 100876, China
% Single address.
% ---------------
\author{Yang Yang, Caili Guo, Fangfang Liu, Chuanhong Liu, Lunan Sun, Qizheng Sun, Jiujiu Chen
\\

\vspace{-0.5cm}
\thanks{
}
}
\vspace{-0.5cm}
%
% For example:
% ------------
%\address{School\\
%   Department\\
%   Address}
%
% Two addresses (uncomment and modify for two-address case).
% ----------------------------------------------------------
%\twoauthors
%  {A. Author-one, B. Author-two\sthanks{Thanks to XYZ agency for funding.}}
%   {School A-B\\
%   Department A-B\\
%   Address A-B}
%  {C. Author-three, D. Author-four\sthanks{The fourth author performed the work
%   while at ...}}
%   {School C-D\\
%   Department C-D\\
%   Address C-D}
%
%\tableofcontents
%\pdfbookmarks
%\ninept
%
\vspace{-0.5cm}
\maketitle
\thispagestyle{empty}

\begin{abstract}
A radical paradigm shift of wireless networks from ``connected things'' to ``connected intelligence'' undergoes, which coincides with the Shanno and Weaver's envisions: Communications will transform from the technical level to the semantic level.
This article proposes a semantic communication method with artificial intelligence tasks (SC-AIT). First, the architecture of SC-AIT is elaborated.
Then, based on the proposed architecture, we implement SC-AIT for a image classifications task. A prototype of SC-AIT is also established for surface defect detection, is conducted. Experimental results show that SC-AIT has much lower bandwidth requirements, and can achieve more than $40\%$ classification accuracy gains compared with the communications at the technical level. Future trends and key challenges for semantic communications are also identified.
\end{abstract}

\vspace{-0cm}
%{\small \emph{Index Terms}---Visible Light Positioning}

\vspace{-0.6cm}
\section{Introduction}
\label{sec:intro}
\IEEEPARstart{W}ith the increasingly powerful intelligence capability, future wireless networks will transform from ``connected things'' to ``connected intelligence'' \cite{Saad2020AVision,Letaief2019The,Xiao2020Toward}. To adapt this new trend,  the intelligent devices in future wireless networks should not only be able to accomplish complex smart applications, but also be utilized for more efficient, robust communications.
Fortunately, this communication paradigm shift coincides with the Shanno and Weaver's envisions for future communications. It is known that conventional wireless networks are based on Shannon's classical information theory. Reviewing Shanno and Weaver's seminal work \cite{Weaver1949Recent}, we can find that communications can be categorized into three levels 1) How accurately can the symbols of communication be transmitted? (The technical problem.); 2) How precisely do the transmitted symbols convey the desired meaning? (The semantic problem.); 3 How effectively does the received meaning affect conduct in the desired way? (The effectiveness problem.).

\begin{figure}[htbp]
\begin{center}
   \includegraphics[width=0.9\linewidth]{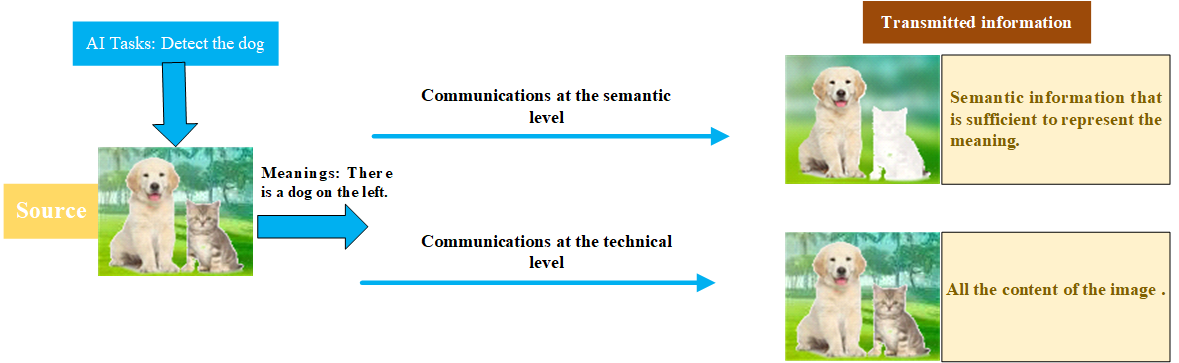}
\end{center}
   \caption{An illustration example of the communications at the semantic level and the technical level.}
\label{fig:example}
\end{figure}

The existing wireless networks limit to the technical level, which aim at transmitting signals as massively and accurately as possible.
In contrast, communications at the semantic level intend to transmit source semantic information.
Here, semantic information can be deemed as the desired ``meaning'' of the source.
For instance, word ``four'' and ``4'' have the same the meaning at the semantic level. However, their representations in bit sequences at the technical level can be totally different. Therefore, semantic communications need to characterize the meanings behind the bits.
Beyond text, the images also contain semantic information.
Fig. \ref{fig:example} shows an example of semantic communications of images, and the target is to detect the dog in the image. Communications at the semantic level only transmit the semantic information related to the dog.
This is because the information is sufficient to represent the meaning of the source, i.e. whether there is a dog in the image.
All the other information related to the background or the cat do not affect the decision-making process, and thus it is not transmitted. However, communications at the technical level treat all information equally important, even though most parts of it do not affect the accomplishment of the artificial intelligence (AI) task. Therefore, communications at the semantic level can significantly improve the communication efficiency \footnote{As stated in Shanno and Weaver's seminal work \cite{Weaver1949Recent}, the effectiveness problem is closely interrelated with the semantic problem, and overlaps it in a rather vague way.
Following this cognition, we generally term the communication method considered in this work as semantic communications. However, it definitely involves the problem in effectiveness level.}.

Even though having promising advantages, there are still some key challenges in semantic communications including the representation and mathematical quantification of semantic information, and the implementation of the semantic communications. The authors in  \cite{Floridi2004Outline} proposed a quantitative theory of semantic information based on the truth-values and semantic discrepancy with respect to a given situation.
The authors in \cite{baoj2011towards} proposed a general model of semantic communications, and the lossless semantic data compression was discussed.
Though the initial theoretical studies such as \cite{Floridi2004Outline,baoj2011towards} shed some light on semantic communications, it is still challenging to implement semantic communications in practice.
In recent years, with the development of machine learning, deep neural networks based semantic communications have been attracting increasing attentions.
The authors in \cite{Xie2021Deep} proposed a semantic communication system for text transmission based on deep learning.
Based on \cite{Xie2021Deep}, the authors in \cite{Xie2021ALite} proposed a lite distributed semantic communication system for internet of thing (IoT) applications.

Even though the existing works are interesting, none of them considered the relations between the AI tasks and the semantic information. Actually, the semantic information is closely related to the target AI task.
For instance, from Fig. \ref{fig:example}, if the target of the transmission changes to detect the cat, the information related to the dog is no longer necessary. Therefore, semantic communication should take AI tasks into considerations.

The main contribution of this paper is a semantic communication paradigm with AI tasks (SC-AIT).
First, we propose the semantic communication architecture for SC-AIT, in which the functions of the three communication levels are clearly identified.
In particular, the effectiveness level mainly concerns with the source information, the desired conduct of the AI tasks, and forms a knowledge base (KB). The semantic level conducts semantic information extraction and coding based on KB and the neural networks. Then, following the design principles of the architecture, we accomplish SC-AIT for a classification task to show its feasibility. Next, an experimental prototype is established to implement SC-AIT for a surface defect detection task, and experimental results verify the efficiency of SC-AIT. We also summarize some open issues and challenges for future semantic communications. Finally, key points of this paper are concluded.

%%%%%%%%%%%%%%%%%%sunlunan%%%%%%%%%%%%%%%%%%%%%%%%%%%%%%%%%%%%
\vspace{-0.3cm}
\section{Architecture and Implementations}
This section elaborates the architecture and implementations of SC-AIT. There are some priori art on the architecture design of semantic communications \cite{Weaver1949Recent,baoj2011towards}. Different from these studies, the proposed architecture introduces AI task block, and investigates the underlaying relations between AI tasks and semantic information to instruct semantic communications.
This is necessary since not all semantic information is required by AI tasks. However, this is also challenging due to the complex relations between the AI tasks and the source information. To circumvent the challenge, this work employs neural networks to achieve the design target. In particular, a study case of SC-AIT is implemented for classification tasks based on an end-to-end neural network. The large number of model parameters in neural networks are utilized to characterize the complex relations between the AI tasks and the source information so as to instruct semantic communications.

\subsection{Architecture Design}
\begin{figure}[htbp]
\begin{center}
   \includegraphics[width=1\linewidth]{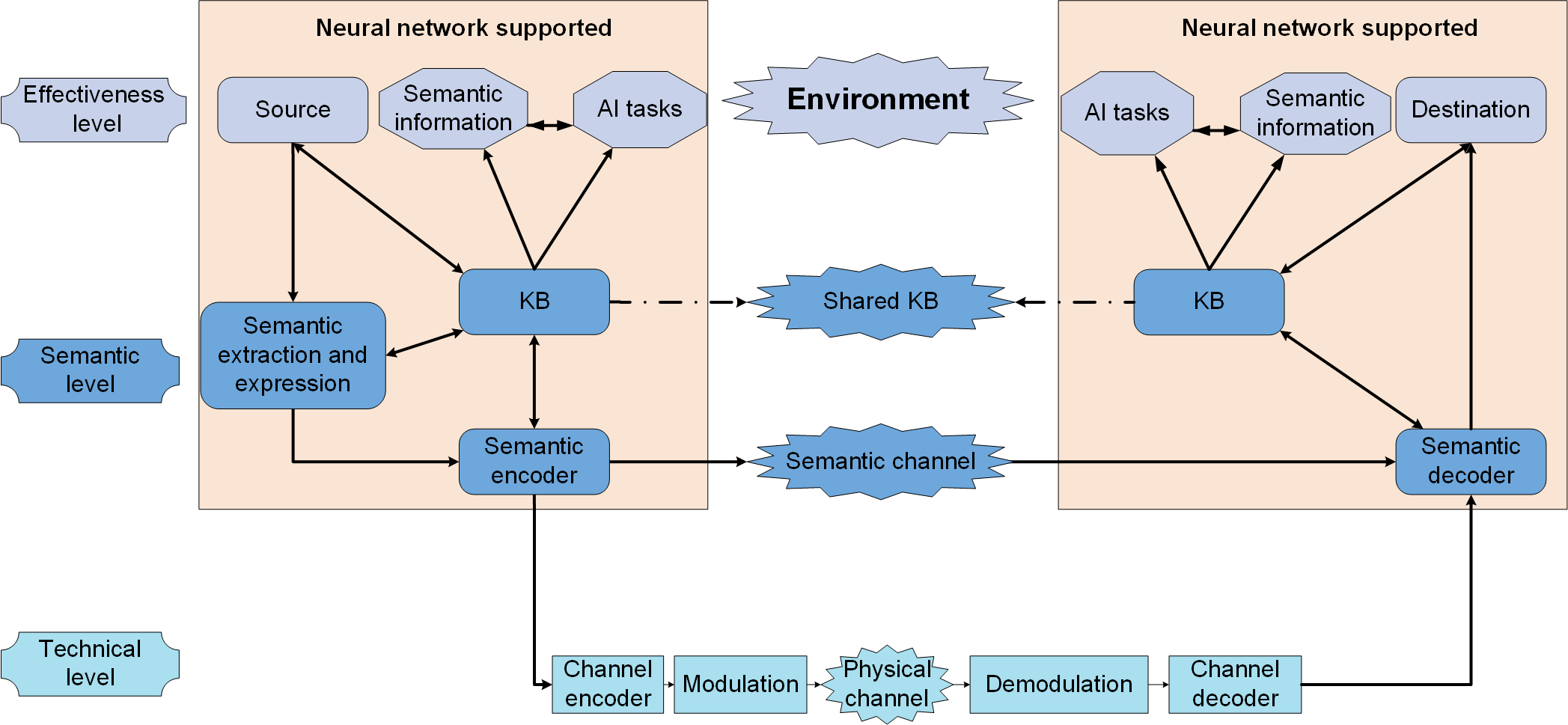}
\end{center}
   \caption{The proposed  semantic communications architecture.}
\label{fig:architecture_en}
\end{figure}

In the first subsection, we introduce the proposed architecture and mainly analyze the necessity and the function of AI task block for the overall semantic communications. The implementation details of the architecture will be elaborated in the next subsection.
As shown in Fig.~\ref{fig:architecture_en}, the semantic communication architecture consists of the effectiveness level, the semantic level and the technical level \cite{Weaver1949Recent}, and each layer is closely interrelated to each other.
At the effectiveness level, there is a source block, which contains the information to be transmitted.
In this work, we assume that the target of the communications is to accomplish a certain AI task that is known by both transmitter and receiver \cite{Letaief2019The}.
The contained information, the AI tasks, and their inherent relations constitute a KB, which can be used to store the above information and instruct the following semantic communications.
Since for different AI tasks, information can have different degrees of importance, it is necessary to introduce AI task in KB.
For instance, as shown in Fig. 1, once the AI task becomes detecting the cat, the information related to dog is no longer necessary.
Moreover, no matter whether the target is the dog or cat, the background information is not needed.
Therefore, the effectiveness level needs to quantify the degrees of importance of semantic information for the accomplishment of AI tasks, so as to decide the parts of semantic information that should be transmitted. At the receiver, there is also a KB to store the knowledge including the related semantic information and AI tasks.

At the semantic level, the semantic information of the source node first needs to be extracted and expressed in a certain form.
After the semantic extraction and expression block, all the semantic information will be delivered to the semantic encoder.
Similar to the conventional source coding, the purpose of the semantic encoder is to reduce source information redundancy.
In particular, SC-AIT proposes to reduce semantic information redundancy according to the contribution to the accomplishment of AI tasks.
The information that is irrelevant to the success of AI tasks can be discarded, and only the useful semantic information is encoded at this block.
Note that the KB has already quantify the relations between the AI tasks and the semantic information, and thus this process can be conducted according to knowledge stored in KB.
At the receiver of the semantic level, a semantic decoder is employed to recover the semantic information.
In addition, there is also a KB at the receiver, which can interact with the source KB via a shared KB at the semantic level.
When the KB at one side alters, the KB at the other side can synchronize the knowledge.
In particular, it is recognized that the destination should be able to increase its level of knowledge by the received message. After the redundancy reduction at the semantic level, the information will be transmitted at the technical level via physical channel.
Note that the local KB can be stored at the transmitter and the receiver, while the shared KB can either be stored in an authoritative third party or just be a virtual KB to synchronize the KBs at the two sides.

At the technical level, the structure is mainly the same as the existing communication systems. One exception is that the source coding is accomplished at the semantic level. This is because the source coding is finished at the semantic encoder block, where the most relevant semantic information is extracted. Compared with the conventional source coding at the technical level, which aims at compressing source information in terms of the amount of the bits, the semantic encoder aims at compressing source information in terms of the content, thus transmitting the ``meaning'' of the source information. Then, channel encoder is employed to enhance the transmission reliability, and the bit stream is modulated for transmission.

\subsection{Implementations}
In this subsection, following the design principles of the proposed architecture, a study case of SC-AIT is implemented for image classification tasks based on convolutional neural networks (CNN). In particular, this subsection elaborates how to establish KB and how to investigate the underlaying relations between AI tasks and semantic information to instruct semantic communications.
%\begin{figure}[htbp]
%\begin{center}
%   \includegraphics[width=1\linewidth]{classification_arch_0607.png}
%\end{center}
%   \caption{Implementations of classification tasks.}
%\label{fig:classification_arc}
%\end{figure}

Image classification is a fundamental AI task that has extensive applications such as surface defect detection in industrial internet, MNIST.
In the considered classification tasks, the source typically contains a number of images, and they will be transmitted to the receiver for analysis.
This is particularly suitable for IoT applications, where the transmitter sensors may have limited computation capability for accurate inferring.
Following the architecture shown in Fig. \ref{fig:architecture_en}, the source image will first be delivered to semantic extraction and expression block to represent the information of the source image. This process can be achieved by a CNN due to its powerful representation learning capability. In particular, a CNN network can extract the key features of the source image and express them into a series of feature maps.
The extracted feature maps indicate different aspects of features of the source images, such as the color, the texture.
Since different AI tasks require different features of the images, it is necessary to further refine the most important features from the extracted features.
This is finished at the semantic encoder block, which further refine the semantic information that is closely related to the AI task.

%Meanwhile, KB of the transmitter collects the involved concepts for the classifications. For instance, for STL 10 dataset, the involved concepts include cat, dog, airplane, etc.. With the target AI tasks and the related concepts, the source KB needs to instruct the key semantic information that should be transmitted.

%
As shown in Fig. \ref{fig:architecture_en}, the semantic encoder block is instructed by source KB. Therefore, before communications, KBs need to be established in advance.
Generally, it is particularly challenging to establish a general KB due to the diversity and the complex relations between the semantic information and the AI tasks.
Fortunately, CNN contains a large number of model parameters. These model parameters are trained not only to minimize a certain loss function for the best AI task performance, but also to determine the best forms of features representing the original image, i.e. semantic information of the source.
Therefore, the model parameters in CNN can naturally characterize the inherent relations between the semantic information and the AI tasks.
In particular, SC-AIT uses the gradients of the output of the CNN with respect to feature maps as the importance weights of the feature map to different classes. Note that the output of the CNN directly decides the class of the image, while the features maps can be deemed as the extracted semantic information in CNN. The gradient actually implies the degrees of importance of semantic information with respect to different classes.
In this way, KB is established, which stores the importance weights of all feature maps for each class. Thus, we quantify the relations between the feature maps (i.e. semantic information) and each class in KBs. Next, the semantic encoder can be conducted based on the established source KB. In particular, we can always transmit the important feature maps by jointly considering the bandwidth and the performance requirements.
Since SC-AIT is achieved in an end-to-end fashion, the receiver KB also requires the knowledge to accomplish the task, and the system needs to synchronize the knowledge once AI tasks or the semantic information changes.
Finally, the compressed semantic information is delivered to channel encoder and modulation for robust transmission.

At the receiver, the destination first demodulates and channel decodes the received message. Fully-connected layers at the receiver act as the semantic decoder. By selecting the class corresponding to the maximum value in the output vector, the destination is able to determine which class the transmitted image belongs to. In this way, SC-AIT is completed for classification tasks.

%\begin{figure}[htbp]
%\begin{center}
%   \includegraphics[width=1.0\linewidth]{STL10_detail_0617.png}
%\end{center}
%   \caption{System blocks of SC-AIT and conventional communications at the technical level.}
%\label{fig:STL10_detail}
%\end{figure}
%
%Fig.~\ref{fig:STL10_detail} illustrates an example of the semantic communications for the class cat.
%First, we extract the features of the source image into 512 feature maps via a CNN networks.
%Then, semantic encoder is implemented based on the established KB, and only the most important feature maps are delivered to the physical transmission link.
%Finally, the receiver can accomplish the classification task with the received semantic information. For comparison, we also illustrate the system blocks of conventional transmission scheme.
%As we can observe, there are two key differences: 1) SC-AIT reduces the source redundancy at the semantic level, while conventional communications achieve this by source coding such as JPEG at the technical level; 2) SC-AIT transmits semantic information and accomplish AI tasks in an end-to-end fashion, while the conventional transmission scheme accomplishes image transmission and classifications in separate steps. Due to these two key differences, it can be expected that SC-AIT will have superior communication efficiency due to the semantic level encoder/decoder, and high reliability to the channel noise due to the end-to-end communication fashion. This will further be verified in the performance evaluation part.

%%%%%%%%%%%%%%%%%%%%%%%%%%%%%%%%%%%%%%%%%%%%%%%%%%%%%%%%%%%%%%%%%%%%%%%%

\vspace{-0.3cm}
\section{Prototype Establishment and Performance Evaluation}
In this section, we establish a semantic communication prototype to implement SC-AIT. In particular, the surface defect detection of  hot-rolled steel strip is considered, which is an image classification task that has been extensively used in industrial IoT to replace subjective and repetitive process of manual inspection.
Then, the performance of SC-AIT is evaluated and compared with baseline schemes based on conventional JPEG transmission and normal semantic communications (SC) without considering AI tasks.

\vspace{-0.3cm}
\subsection{Proof-Of-Concept Prototype}
\begin{figure}[htbp]
	\begin{center}
		\includegraphics[width=0.9\linewidth,height=0.5\textheight]{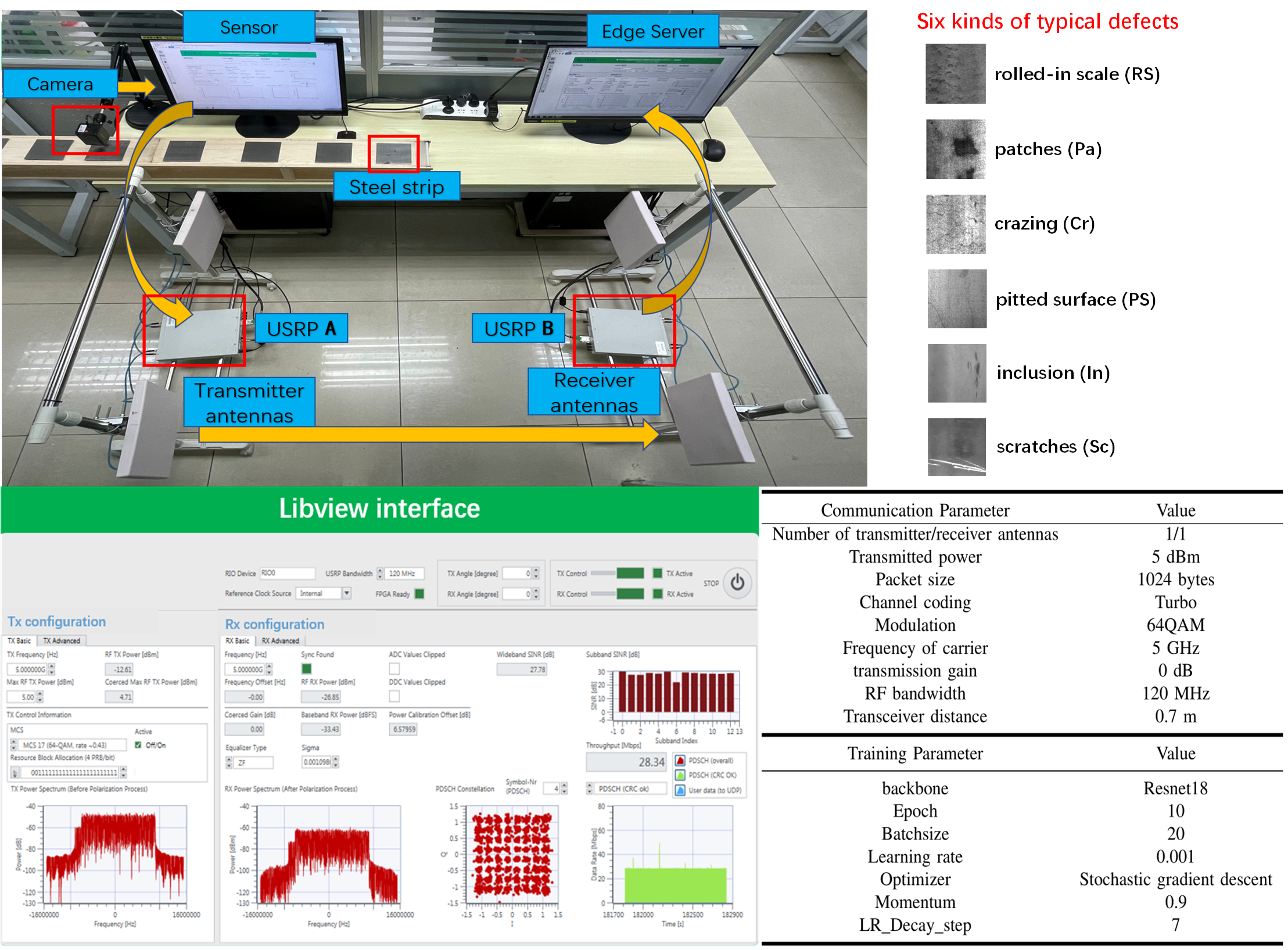}
	\end{center}
	\caption{Semantic communication prototype for surface defect detection tasks.}
	\label{fig:system}
\end{figure}
Fig.~\ref{fig:system} illustrates the established prototype. As shown in Fig.~\ref{fig:system}, we assume the transmitter is a sensor used to inspect surface defects. In particular, the transmitter is composed of a camera, a universal software radio peripheral (USRP), transmitter antennas.
The camera acts as a source to capture the images of the hot-rolled steel strip for surface defects. Then, the images from the camera are delivered to the sensor. Here, sensor acts as an intelligent node used to inspect surface defects, and its function is achieved by a computer in our prototype. In particular, the sensor first extracts and codes the semantic information of the images according to the established source KB, and then modulates the signals at the technical level. The USRP at the transmitter side (USRP A) reads the byte streams from the User Datagram Protocol (UDP) port of the sensor via Peripheral Component Interconnect express (PCIe) bus. USRP is software-defined radios, which conducts digital to analogue conversions and adapts signals to suitable forms for transmission. The LabVIEW software is used to control the transmission parameters, such as the frequency of the carrier and the transmission gains, as shown in Fig.~\ref{fig:system}. Finally, the antennas will transmit the signals.

The receiver is composed of receiver antennas, a USRP, and an edge server acted by a computer at the receiver side. In particular, the receiver antennas first deliver the received signal to the USRP at the receiver side (USRP B). In USRP B, the signals representing the semantic information of the image are demodulated and decoded into byte streams. Then, the byte streams are delivered to edge server's UDP port via PCIe bus. The edge node then decodes the semantic information and outputs the class that semantic information indicates.

The key system parameters are summarized in Fig. \ref{fig:system}. The dataset is NEU surface defect dataset\footnote{http://faculty.neu.edu.cn/yunhyan/NEU\_surface\_defect\_database.html}, which is suitable for industrial Internet scenarios. As shown in Fig.~\ref{fig:system}, six kinds of typical surface defects of the hot-rolled steel strip are collected in the dataset, i.e., rolled-in scale, patches, crazing, pitted surface, inclusion and scratches. The dataset includes 1800 grayscale images (300 samples for each kind), and the original resolution of each image is $200 \times 200$ pixels.

%\begin{table}[htbp]
%	\small
%	\centering
%	\caption{System Parameters.}
%	\setlength{\abovecaptionskip}{-0.5cm}
%	\begin{tabular}{ccc}
%		\toprule[2pt]
%		Communication Parameter & Value \\
%		\hline
%		Number of transmitter/receiver antennas & 1/1 \\
%        Transmitted power & 5 dBm \\
%		Packet size & 1024 bytes\\
%		Channel coding & Turbo\\
%		Modulation & 64QAM\\
%		Frequency of carrier & 5 GHz\\
%		transmission gain & 0 dB\\
%		RF bandwidth & 120 MHz\\
%		Transceiver distance & 0.7 m\\
%		\midrule[2pt]
%		Training Parameter & Value \\
%		\midrule
%        backbone & Resnet18 \\
%		Epoch & 10 \\
%		Batchsize & 20\\
%		Learning rate & 0.001\\
%		Optimizer & Stochastic gradient descent\\
%		Momentum & 0.9\\
%		LR\_Decay\_step & 7\\
%		\bottomrule[2pt]
%	\end{tabular}
%	\vspace{-0.cm}
%\end{table}

\subsection{Performance Evaluations}
This subsection evaluates the performance of the proposed SC-AIT. A conventional scheme that directly transmits images in JPEG forms for classifications is employed as the baseline scheme. In addition, the proposed communication method under different semantic compression ratios (CR) are considered. For instance, $CR=0\%$ indicates that all the feature maps are transmitted without compression, while $CR=30\%$ indicates 30\% of the feature maps are discarded, while the remaining 70\% feature maps are transmitted.

\begin{figure}[htbp]
	\begin{center}
		\includegraphics[width=0.8\linewidth]{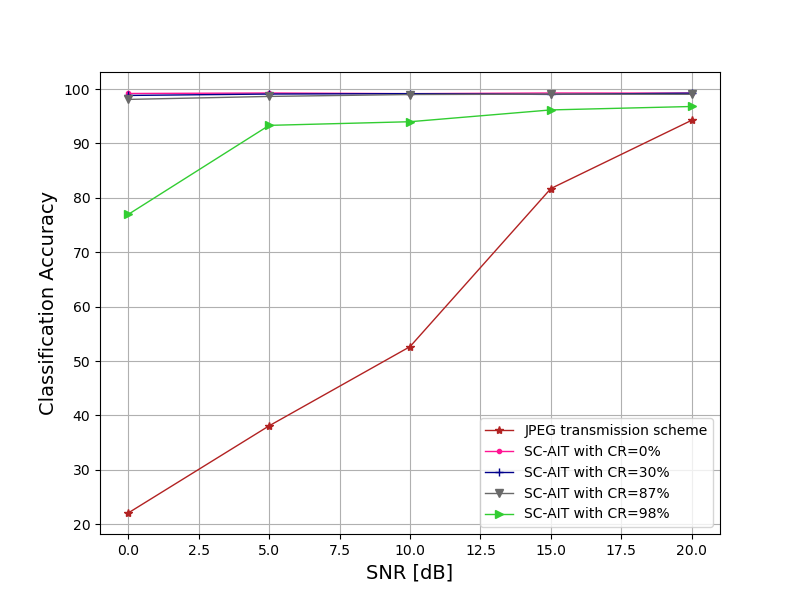}
	\end{center}
	\caption{Simulation results of classification accuracy under different SNRs.}
	\label{fig:simulation2}
\end{figure}
Fig.~\ref{fig:simulation2} first compares the classification accuracy of SC-AIT and the conventional JPEG transmission scheme under different signal to noise ratios (SNRs). As shown in Fig.~\ref{fig:simulation2}, compared with JPEG compressed scheme, the semantic communication can always obtain better performance.
In particular, SC-AIT with $CR=98\%$ can still harvest more than 40\% accuracy gains even at an aggressive compression ratio and 10 dB SNR when compared with the JPEG transmission scheme, since the top 2\% most relevant semantic information is transmitted.
We can also observe that the performance gain is more significant at low SNR regions. This is because the semantic communication system conducts in an end-to-end fashion and takes channel state information into the training. In contrast, the conventional JPEG transmission scheme has much worse bit error rate at low SNRs, leading to worse classification performance. Moreover, when $CR = 87\%$, there is almost no loss of classification accuracy when SNR varies. This indicates that SC-AIT can significantly reduce the redundant information without affecting the task performance.

\begin{figure}[htbp]
	\begin{center}
		\includegraphics[width=0.8\linewidth]{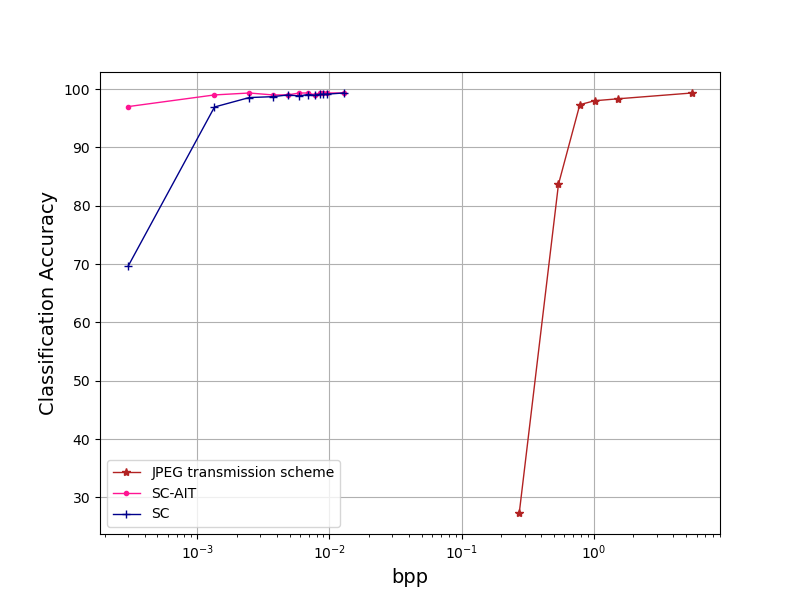}
	\end{center}
	\caption{Experimental results of classification accuracy with different available bandwidths.}
	\label{fig:simulation1}
\end{figure}
Fig.~\ref{fig:simulation1} compares the performance of the proposed SC-AIT with a conventional JPEG transmission scheme and a normal semantic communication method without considering AI tasks (SC) under different bits per pixel (bpp). Here, different bpps indicates different bandwidth requirements of the communication methods. The lower the bpp is, the more extreme the source has to be compressed, and thus less bandwidth is required. In addition, since normal SC does not consider AI task, it tries to extract and transmit all the features. When the bpp constraint does not support transmitting all the feature maps, SC will randomly transmits parts of feature maps, as it does not consider the relations between the AI tasks and feature maps. In contrast, the proposed SC-AIT always transmits the most important parts of feature maps to satisfy different bpp constraints.
As we can observe from Fig.~\ref{fig:simulation1}, both SC-AIT and SC can achieve much lower bpps when compared with conventional JPEG transmission scheme. In particular, both the schemes can achieve convergent performance at $10^{-2}$ level of bpp.
Moreover, we can also observe that SC-AIT can achieve more significant performance gains when bpp is extremely low (lower than $10^{-3}$). This is because SC-AIT chooses the most important feature maps for transmission, which are the semantic information closely related to the success of AI tasks.

\begin{figure}[htbp]
	\begin{center}
		\includegraphics[width=0.8\linewidth]{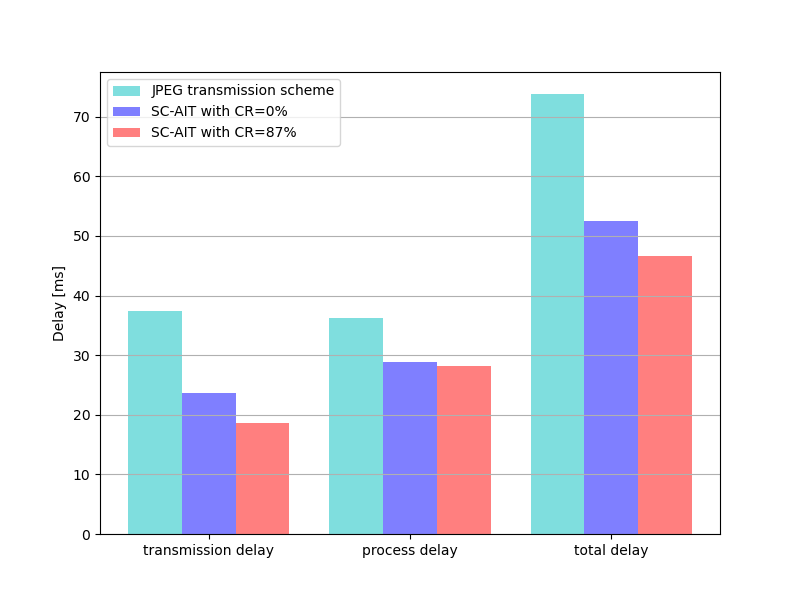}
	\end{center}
	\caption{Experimental results of complexity in terms of runtime.}
	\label{fig:delay}
\end{figure}
Fig.~\ref{fig:delay} compares the complexity of SC-AIT and the baseline schemes in terms of the system runtime. In particular, the runtime includes transmission delay and process delay. The computers used for experiments are equipped with Intel(R) Xeon(R) Gold 6240 CPU @ 2.60GHz, 100G RAM, and Tesla V100 32G graphics card.
As shown in Fig.~\ref{fig:delay}, the proposed SC-AIT have both lower process delay and transmission delay compared with the conventional JPEG transmissions.
The total delay of SC-AIT is only 70\% of the conventional JPEG transmission scheme.
This is because the SC-AIT uses the feature extraction network to extract the semantic information of the image, which has lower complexity than that of the conventional, complex source coding. The total complexity is thus reduced. When the feature maps are further compressed by 87\%, the transmission delay is further reduced by 21\%.
This is because the transmitted information is further compressed. Therefore, we can conclude that the semantic communication system has low complexity and is particularly suitable for delay-sensitive applications and systems with limited computation capability.

%%%%%%%%%%%%%%%%%%%%%%%%%sunqizheng%%%%%%%%%%%%%%%%%%%%%%%%%%%%%%%%%%%%%%%%%%%%%%%
\vspace{-0.2cm}
\section{Challenges and Opportunities}
The above work can serve as basis for semantic communications that takes AI tasks into consideration. This section further summarizes some key challenges  remaining in the research field, and sheds some light on future research opportunities.

\subsection{Mathematical Fundamentals of Semantic Communications}
It is known that conventional wireless networks are based on Shannon's classical information theory, in which basic definitions such the amount of information and channel capacity are rigorously derived. Similarly, semantic communications also require such basic definitions to support. There are a series of key issues to be determined: How to quantify semantic information? What is the lossless semantic compression? What is the best representation of semantic information?
These basic questions are essential to derive the theoretical semantic communication capacity bound. There are several initial works that discussed this topic. For instance, in \cite{baoj2011towards}, the authors analyzed the lossless semantic compression and establish theoretical bounds in lossless semantic data compression. However, the work has non-trivial assumptions such as adopting the logic-based approaches to capture semantic in human communication, which can significantly limit the application of the bound and utility communication in the future. In addition, general theories such as the information bottleneck theory may be applied. The information bottleneck theory extends the rate-distortion theory of data compression, and is able to explain the deep learning process. In particular, the information bottleneck theory can be used to express the source information into a briefest form, which may used to instruct the semantic compression and communications. In conclusion, the mathematical fundamentals of semantic communications is the basis of semantic communications. It is vital to quantify, express and derive the lossless bound of semantic communications.

\subsection{Neural Network Structure Optimization}
In our implementations of SC-AIT, we extract the semantic information based on a given network. Actually, the neural network structure is also closely related to the extracted semantic information.
For instance, in \cite{net2vec2018} , the Net2Vec framework is proposed, which maps semantic information to the corresponding feature filter space, and analyzes how semantic information is encoded by CNN filters.
In \cite{gradcam2017}, Grad-CAM, a class-discriminative localization technique is proposed, which analyzes the importance of feature maps to concepts, and generates visual explanations for the CNN network architecture.
Both works verify that the semantic features extracted by CNN is closely related to the structure of the CNN.
Therefore, the neural network structure needs to be optimized according to the information in the KB, so that CNN can represent semantic information more accurately.
Furthermore, the loss function of a neural network can also affect the extracted semantic information.
A appropriate loss function can make feature extracted have better task performance \cite{perceptualloss2016}.
Therefore, the neural network structure and the loss function needs to be optimized to represent semantic information more well while at moderate cost.

\subsection{C4 Resource Allocation}
Semantic communications require seamless integration of communication and intelligence. Therefore, besides communication resources, computation resource, caching resource, and control strategies (C4) are also deeply involved in semantic communications. Moreover, the C4 resources are closely interacted\cite{3C2017}, and all of them can significantly affect the final AI task performance. Therefore, the C4 resource allocation plays a key role in semantic allocation. It is challenging to consider the joint allocation of 4C resources according to the mutual constraint and restriction relationship of 4C resources. Besides, the resource allocation rule may also alter in semantic communications. In conventional wireless networks, the resources are typically allocated according to the QoS or QoE of users. In contrast, in semantic communications, the C4 resources need to be allocated according to the importance of the semantic information. Therefore, the resource allocation in semantic communications also requires new rules or principles.

%\subsection{Task Extensions}
%This work mainly implements semantic communications for image classification tasks to verify the feasibility and efficiency. SC-AIT should also be able to applied to extensive applications such as object localization, semantic segmentation, and instance segmentation. There are still several key challenges. In particular, the relations between the semantic information and AI tasks can be more complex in these AI tasks. In addition, the neural networks for more complex AI tasks can also be more complex. The performance evaluation indexes of complex tasks are diversified. It is significant that how to select and use these indexes as the target of semantic extraction. In general, for different tasks, a variety of AI methods should be exploited to assist semantic communication in the future, and the future semantic communications system should apply to all kinds of AI tasks.

\section{Conclusion}
This work has proposed a semantic communication method that takes AI tasks into consideration. The architecture of the proposed SC-AIT has been introduced, which is compatible with the Shanno and Weaver's envision for semantic communications. The function of each block in the proposed architecture has also been introduced in detail. Then, we have elaborated the proposed SC-AIT for classification tasks. A prototype is established for SC-AIT with surface defect detection tasks. Experimental results have verified the efficiency of the proposed semantic communications. We have also shed some light for the future researchers.


\begin{thebibliography}{10}

\bibitem{Letaief2019The}
K.~B.~Letaief, W.~Chen, Y.~Shi, J.~Zhang, and Y.~A.~Zhang,
\newblock ``The Roadmap to 6G: AI Empowered Wireless Networks,''
\newblock {IEEE Commun. Mag., vol. 57, no. 8, pp. 84-90, Aug. 2019.}

\bibitem{Saad2020AVision}
W.~Saad, M.~Bennis, and M.~Chen,
\newblock ``A Vision of 6G Wireless Systems: Applications, Trends, Technologies, and Open Research Problems,''
\newblock {IEEE Netw., vol. 34, no. 3, pp. 134-142, Jun. 2020.}

\bibitem{Xiao2020Toward}
Y.~Xiao, G.~Shi, Y.~Li, W.~Saad, and H.~V.~Poor,
\newblock ``Toward Self-Learning Edge Intelligence in 6G,''
\newblock {IEEE Commun. Mag., vol. 58, no. 12, pp. 34-40, Dec. 2020.}

\bibitem{Weaver1949Recent}
W.~Weaver,
\newblock ``Recent Contributions to The Mathematical Theory of Communication,''
\newblock {ETC: a review of general semantics, vol. 1, no. 1, pp. 261-281, Sep. 1949.}

\bibitem{Carnap1954Outline}
R.~Carnap, and Y. Bar-Hillel,
\newblock ``An Outline of a Theory of Semantic Information,''
\newblock {Technical Report, no. 247, Oct. 1952.}

\bibitem{Floridi2004Outline}
L.~Floridi,
\newblock ``Outline of a Theory of Strongly Semantic Information,''
\newblock {Minds and Machines, vol. 14, no. 2, pp. 197-221, 2004.}

%\bibitem{Selvaraju_2017}
%Selvaraju R R, Cogswell M, Das A, et al.
%\newblock ``Grad-cam: Visual explanations from deep networks via gradient-based localization''
%\newblock {Proceedings of the IEEE international conference on computer vision. 2017: 618-626.}
%%%%%%%%%%%%%%%%%%%%%%%sunqizheng%%%%%%%%%%%%%%%%%%%%%%%%%%%%%%%%%%%%%%%%%%%%%%%%%
\bibitem{baoj2011towards}
J. Bao, P. Basu, and M. Dean,
\newblock ``Towards a theory of semantic communication,''
\newblock {IEEE Network Science Workshop, New York, USA, Jun. 2011.}

\bibitem{Xie2021Deep}
H. Xie, Z. Qin, G. Y. Li, and B. Juang,
\newblock ``Deep Learning Enabled Semantic Communication Systems,''
\newblock {IEEE Trans. Signal Process., vol. 69, no. 1, pp. 2663-2675, Apr. 2021.}

\bibitem{Xie2021ALite}
H. Xie, and Z. Qin,
\newblock ``A Lite Distributed Semantic Communication System for Internet of Things,''
\newblock {IEEE J. Sel. Areas Commun., vol. 39, no. 1, pp. 142-153, Jan. 2021.}

\bibitem{net2vec2018}
R. Fong, and A. Vedaldi,
\newblock ``Net2vec: Quantifying and explaining how concepts are encoded by filters in deep neural networks,''
\newblock {in Proc. of the IEEE conference on computer vision and pattern recognition, Salt Lake City, Utah., USA, 2018.}

\bibitem{gradcam2017}
R. Selvaraju, M. Cogswell, and A. Das,
\newblock ``Grad-cam: Visual explanations from deep networks via gradient-based localization,''
\newblock {in Proc. of the IEEE international conference on computer vision, Venice, Italy, 2017.}

\bibitem{perceptualloss2016}
J. Johnson, A. Alahi, and F. Li,
\newblock ``Perceptual losses for real-time style transfer and super-resolution,''
\newblock {in Proc. of the European conference on computer vision, Amsterdam, Netherlands, 2016.}

\bibitem{3C2017}
Y. Zhou, F. Yu, and J. Chen,
\newblock ``Resource allocation for information-centric virtualized heterogeneous networks with in-network caching and mobile edge computing,''
\newblock {IEEE Trans. Veh. Technol., vol. 66, no. 12, pp. 11339-11361, Aug. 2017.}

\end{thebibliography}
\end{document}